\documentclass[letterpaper]{article} 
\usepackage{aaai23}  
\usepackage{times}  
\usepackage{helvet}  
\usepackage{courier}  
\usepackage[hyphens]{url}  
\usepackage{graphicx} 
\usepackage{natbib}  
\usepackage{caption} 
\usepackage{algorithm}
\usepackage{algorithmic}

\newcommand{\ie}{\textit{i}.\textit{e}.}

\usepackage{amsmath}
\usepackage{amssymb}
\usepackage{makecell}
\usepackage{multirow}
\usepackage{booktabs}
\usepackage{arydshln}

\usepackage{newfloat}
\usepackage{listings}
\DeclareCaptionStyle{ruled}{labelfont=normalfont,labelsep=colon,strut=off} 
\floatstyle{ruled}
\newfloat{listing}{tb}{lst}{}
\floatname{listing}{Listing}
\title{Deep Visual Forced Alignment:\\Learning to Align Transcription with Talking Face Video}
\author{
    Minsu Kim, Chae Won Kim, Yong Man Ro\thanks{Corresponding author.}
}
\affiliations{
    Image and Video Systems Lab, KAIST, South Korea\\
    \{ms.k, chaewonkim, ymro\}@kaist.ac.kr
}

\usepackage{bibentry}

\begin{document}

\maketitle

\begin{abstract}
Forced alignment refers to a technology that time-aligns a given transcription with a corresponding speech. However, as the forced alignment technologies have developed using speech audio, they might fail in alignment when the input speech audio is noise-corrupted or is not accessible. We focus on that there is another component that the speech can be inferred from, the speech video (\ie, talking face video). Since the drawbacks of audio-based forced alignment can be complemented using the visual information when the audio signal is under poor condition, we try to develop a novel video-based forced alignment method. However, different from audio forced alignment, it is challenging to develop a reliable visual forced alignment technology for the following two reasons: 1) Visual Speech Recognition (VSR) has a much lower performance compared to audio-based Automatic Speech Recognition (ASR), and 2) the translation from text to video is not reliable, so the method typically used for building audio forced alignment cannot be utilized in developing visual forced alignment. In order to alleviate these challenges, in this paper, we propose a new method that is appropriate for visual forced alignment, namely Deep Visual Forced Alignment (DVFA). The proposed DVFA can align the input transcription (\ie, sentence) with the talking face video without accessing the speech audio. Moreover, by augmenting the alignment task with anomaly case detection, DVFA can detect mismatches between the input transcription and the input video while performing the alignment. Therefore, we can robustly align the text with the talking face video even if there exist error words in the text. Through extensive experiments, we show the effectiveness of the proposed DVFA not only in the alignment task but also in interpreting the outputs of VSR models.
\end{abstract}

\section{Introduction}
When we watch a foreign movie, we naturally expect time-aligned subtitles to appear with the dialogue presented in the film. However, manually annotating the time-alignment information between the transcription and speech is burdensome. To reduce the burden, we can automatically generate the time-aligned subtitles by using a technique called forced alignment. Forced alignment aims to align the given transcription with the corresponding speech and produce the timeline information in word- or phoneme-levels for the transcription. Methods for implementing forced alignment can be categorized into two main approaches. The first approach is utilizing audio-based Automatic Speech Recognition (ASR) methods that utilize Hidden Markov Models (HMM) \cite{rabiner1989HMM} or Connectionst Temporal Classification (CTC) \cite{graves2006CTC}. By tracking the states of HMM \cite{mcauliffe2017MFA} or finding the CTC-segment \cite{kurzinger2020ctc-align}, we can obtain the time-alignment of a given transcription. The second approach is utilizing Dynamic Time Warping (DTW) \cite{muller2007DTW}, a time series alignment algorithm. The transcription is firstly converted into audio with timelines using Text-to-Speech (TTS). Then, by using DTW, the synthesized audio is aligned with the real audio to produce the alignment map \cite{alberto2017aeneas}. Since forced alignment technologies generally assume that the audio is accessible and clean, they could fail to align the transcription with speech when the audio is recorded in a noisy environment or is absent.

\begin{figure}[t!]
	\begin{minipage}[b]{1.0\linewidth}
		\centering
		\centerline{\includegraphics[width=8.6cm]{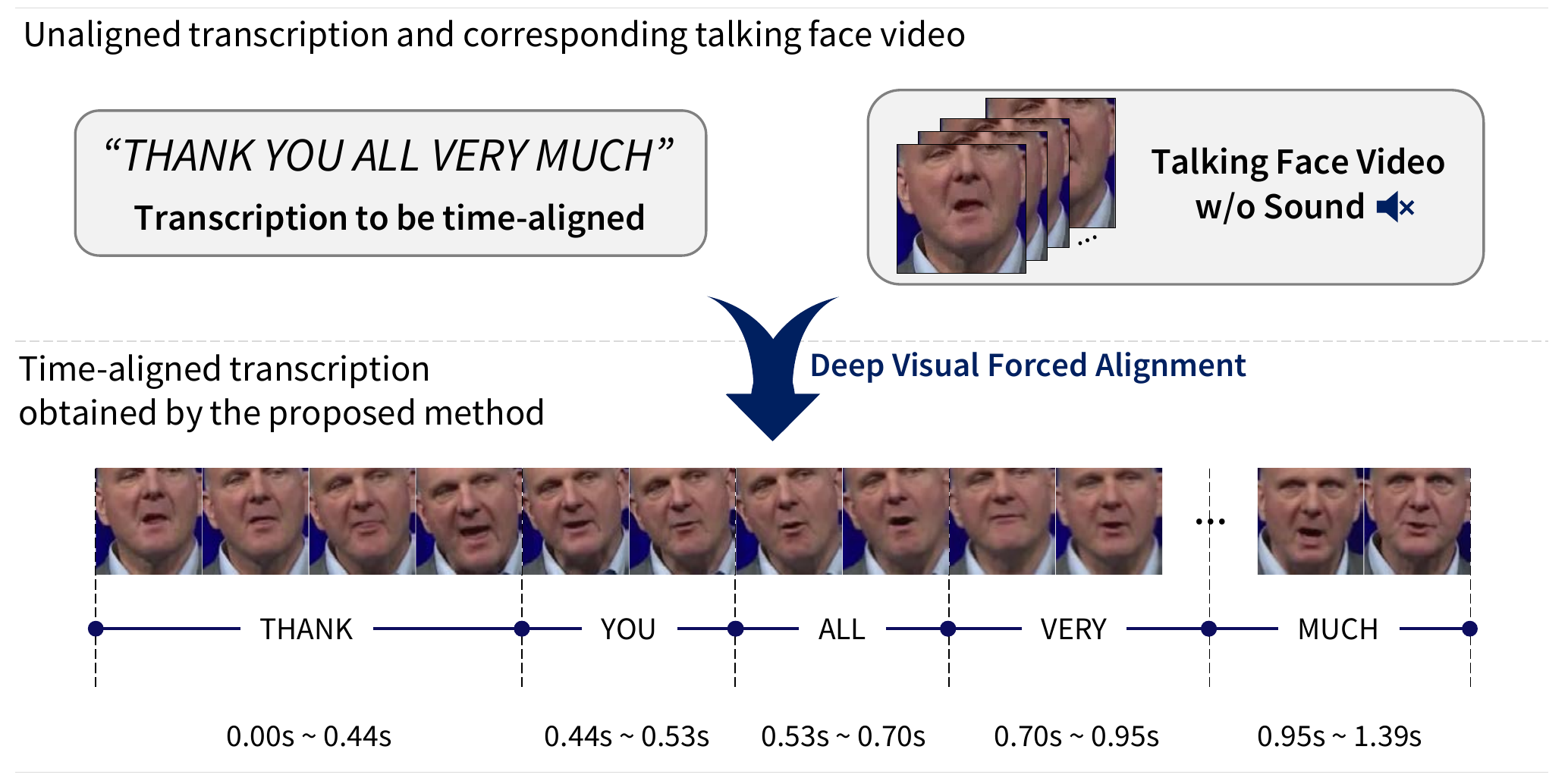}}
	\end{minipage}
	\caption{The proposed DVFA can time-align a transcription with a talking face video by watching its lip movements.}
	\label{fig:1}
\end{figure}

These days, Visual Speech Recognition (VSR) has drawn big attention for its complementary effect that it can assist or replace ASR in noisy environments or even when the audio is absent. This is possible since lips participate in producing speech, and thus lip movements and speech audio are highly correlated \cite{sataloff1992humanvoice}. Analogous to that VSR developed as a complementary for ASR, one can try to design a video-based forced alignment technology (denoted as visual forced alignment) to complement the audio-based forced alignment technology (denoted as audio forced alignment) in poor-audio conditions. However, different from audio forced alignment, developing a reliable visual forced alignment is challenging due to the following two reasons. Firstly, the performance of VSR is still not enough to trust the prediction results. The state-of-the-art VSR method achieves about 27\% Word Error Rate (WER) while ASR method easily earns about 1\% WER on the same dataset \cite{shi2022avhubert}. This performance gap between VSR and ASR makes it hard to develop a visual forced alignment method using recognition-based approaches as done in audio forced alignment (\ie, using HMM or CTC). Secondly, the reverse directional translation, text to talking face generation, has also inferior performance to that of TTS. The generated lip movements fail to contain the true pronunciations and instead look like puppetry \cite{agarwal2020detecting}. Therefore, using DTW to align the text with video could produce unexpected results. In order to align the transcription with a given talking face video without depending on the sound, a new method that appropriates for visual forced alignment should be considered.

In this paper, we propose a Deep Neural Network (DNN) based visual forced alignment method, namely Deep Visual Forced Alignment (DVFA), a novel method that can align given transcriptions with talking face videos by only watching its lip movements as shown in Fig. \ref{fig:1}. The proposed DVFA utilizes multi-modal attention to match transcriptions with the corresponding lip movement frames at word level. Through the multi-modal attention, the corresponding frames to a given word can be retrieved. Moreover, we augment the alignment task with anomaly case detection which can distinguish whether the given words in the transcription actually match well with the speech derived from the talking face video. The effectiveness of the proposed DVFA is evaluated on popular benchmark databases, LRS2 \cite{chung2017lrs2} and LRS3 \cite{afouras2018lrs3}, and we show that the proposed DVFA can be employed not only for forced alignment but also for validating the outputs of VSR to boost its practicality.

Our contributions can be summarized as follows:
\begin{itemize}
    \item To the best of our knowledge, this is the first work that tries to align the sentence-level transcription with talking face video without using the speech audio.
    \item By augmenting the alignment task with anomaly case detection which is for distinguishing whether the word in transcription is actually spoken by the person in the video, the proposed DVFA can perform forced alignment robust to errors in transcription, and can be utilized to enhance the practicality of VSR technology.
    \item The proposed DVFA achieves state-of-the-art alignment performances on LRS2 and LRS3 databases compared to CTC-based method and the state-of-the-art visual keyword spotting methods. Moreover, through qualitative and quantitative analysis, we verify the effectiveness of the proposed method in both forced alignment, and as an interpreter for VSR systems.
\end{itemize}

\section{Related Work}
\subsection{Audio Forced Alignment}
Forced alignment is the task of finding the timeline of each word or phoneme in transcriptions to match with the speech. It is very useful for diverse applications. For example, it can be used to add subtitles for movies and TV programs, and to develop TTS models \cite{ren2020fastspeech2,elias2021parallel} which require aligning the speech with the text in advance to be trained. Forced alignment methods are mainly developed by using speech audio \cite{yuan2008speaker,gorman2011prosodylab}. \citet{mcauliffe2017MFA} proposed a forced alignment method by utilizing HMM-based speech recognition. By tracking the state of HMM, it can get the time-alignment map between transcription and speech. \citet{kurzinger2020ctc-align} proposed to align the text with speech by using CTC-based speech recognition \cite{graves2006CTC}. By finding the best path for the recognition results, they can obtain the CTC-segment which holds the timeline information of transcription matched to the speech. \citet{alberto2017aeneas} employed DTW algorithm \cite{muller2007DTW} to perform forced alignment. By applying DTW between synthesized audio from TTS and the real audio, they can align the transcription with the speech. Recently, \citet{li2022neufa} proposed a deep learning-based audio forced alignment method. The aforementioned audio forced alignment methods are developed under the assumption that the speech audio exists and is not corrupted with severe noise. Therefore, naturally, we could not get the text-speech alignment when the audio is not accessible or is recorded with various environmental noises. Different from the aforementioned works, we try to develop a method to align the transcription with a talking face video without using speech audio. Since the video has different characteristics from the audio such as lower speech recognition performance and inferior text-to-lip translation performance, it is not an easy problem to develop video-based forced alignment technology. To handle this, we propose a novel DNN-based Deep Visual Forced Alignment (DVFA) that can accurately align the input transcription with the input talking face video.

\begin{figure*}[t!]
	\begin{minipage}[b]{1.0\linewidth}
		\centering
		\centerline{\includegraphics[width=18cm]{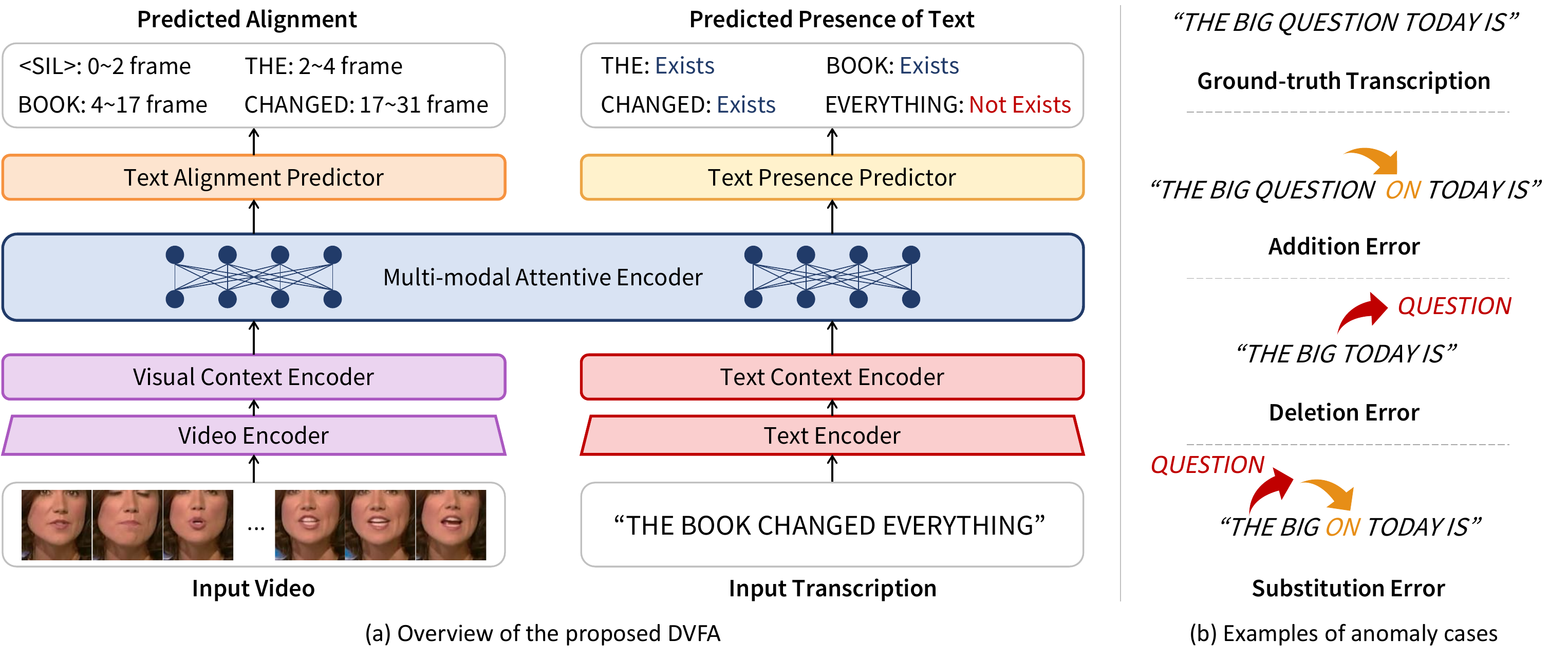}}
	\end{minipage}
	\caption{(a) The overall architecture of the proposed DVFA. It can predict the time-alignment between the transcription and the video, and whether there exist anomaly errors. (b) The anomaly cases (\ie, addition, deletion, and substitution error).}
	\label{fig:2}
\end{figure*}

\subsection{Visual Keyword Spotting and Video Localization}
Visual Keyword Spotting (KWS) is a task of detecting and localizing keywords in an input sequence of silent video. KWS is practical in situations when there is a lack of or noisy audio, or when it is not necessary to detect every word like in wake-word recognition. Existing methods proposed different model systems to improve KWS performance on various audio-visual speech datasets. \citet{yao2019spotting} used a sliding windows strategy to extract sentence-level embeddings and obtain the occurence probability of keywords. \citet{jha2018word} introduced a query by exemplar method, which uses video as both the query and retrieval to compute the correlation between spatio-temporal features of the query and retrieved candidates. \citet{stafylakis2018zero} used a grapheme-to-phoneme encoder-decoder architecture to model words to their pronunciation. Recent works \cite{momeni2020KWS, prajwal2021transpotter} utilized a CNN and similarity map and the transformer \cite{vaswani2017attention} with cross-modal attention to align phonetic and visual features, respectively. However, since KWS is developed to localize one word in a video, it is prone to produce overlapped time-alignment map between different words when it is applied to align the whole sentences with the video. Moreover, when the same word appears multiple times in the sentences, it could fail in localizing all words simultaneously and could localize just one word among them. Different from KWS methods, we propose a method aligning the whole sentences with the given talking face video. Moreover, not only producing the alignment map, but the proposed method can also detect mismatches including addition, deletion, and substitution errors, between the text and the talking face video.

Another related research is temporal localization in video using text queries, which tries to find a video segment that the query sentence describes \cite{bojanowski2015local,gao2017local,anne2017local,yuan2019local,lee2022audio}. Different from these works, our work tries to time-align the transcription with the face video that saying the sentence in the transcription. Thus, while the aforementioned study focuses on semantic modeling between sentences and videos, our objective is to track lip movements and infer the mentioned words for accurate time alignment.

\subsection{Visual Speech Recognition}
Visual Speech Recognition (VSR) is a technology predicting speech by solely depending on video inputs. The technology has largely progressed in both network architectural aspects \cite{stafylakis2017reslstm, martinez2020mstcn, assael2016lipnet, chung2017lrs2, afouras2018deep, ma2021confavsr} and training strategies \cite{afouras2020asr, zhao2020hearing, ren2021learning, kim2021multi, kim2022distinguishing, kim2022speaker}. Even though the performance of VSR has improved for decades, it is still not enough to be utilized in practice. The state-of-the-art VSR achieves performance over 25\% WER, which is a very high error rate compared to that of ASR. On the same dataset, ASR can easily obtain under 2\% WER \cite{shi2022avhubert}. In this paper, we introduce a method to improve the practicality of VSR by using the proposed DVFA as an interpreter for the outputs of VSR models.

\section{Proposed Method}
Let $x_v \in \mathbb{R}^{T \times H \times W \times C}$ be a talking face video with frame lengths of $T$, frame size of $H \times W$, and channel size of $C$, and $x_t \in \mathbb{R}^{S}$ be a corresponding transcription to the talking face video said, where $S$ represents the length of transcription. Our objectives are 1) finding time-alignment information of the input transcription which matches with the input video and 2) detecting anomaly cases of whether the input transcription is well matching with the input talking face video. To time-align the words with the talking face video, we design a multi-modal attentive encoder that can attend across the two modalities, text and video. With the multi-modal attentive encoder, the corresponding video frames saying a given word can be retrieved, and thus it is able to align the word at the frame-level. 
Moreover, to detect anomaly errors, we consider three anomaly cases, addition error, deletion error, and substitution error (shown in Fig. \ref{fig:2}(b)). 
To handle the anomaly cases, we design two output predictors. The first is Text Alignment Predictor (TAP) which predicts the word alignment information for each input frame, and the second is Text Presence Predictor (TPP) which predicts the presence of text in the video for each input text token. Therefore, with the proposed method, we are able to not only align the given transcription with the talking face video without accessing audio, but also find the anomaly errors, \ie, addition and deletion errors. These properties of the proposed method give us advantages of robustness to the mismatches between input transcription and input video in forced alignment, and usefulness in verifying the outputs of a pre-trained VSR model in real-world applications. The proposed DVFA is illustrated in Fig. \ref{fig:2}(a).

\subsection{Modality-specific Representation Learning}
In order to align the two different modal representations, visual and text, they are firstly embedded by respective modality-specific encoders, $E_v$ and $E_t$. To model the intra-modality context which is proven as keys in visual \cite{kim2021lip2speech} and text \cite{devlin2018bert} representation learning, we employ conformer \cite{gulati2020conformer} which can capture both global and local information of input sequences. Let $C_v$ and $C_t$ be the visual context encoder and text context encoder, respectively. Then, the intra-modality context for each modality can be modeled as follows,
\begin{align}
    f_v=C_v(E_v(x_v)), \quad f_t=C_t(E_t(x_t)),
\end{align}
where $f_v\in\mathbb{R}^{T\times D}$ is the encoded visual context and $f_t\in\mathbb{R}^{S\times D}$ is the encoded text context with embedding size of $D$. With the obtained visual and text context representations, we try to model the inter-modal relationships to find the time-alignment map between the two modalities.

\subsection{Multi-modal Attentive Learning}
When people try to find the alignment of given words with a talking face video, they naturally skim through the whole video frames and attend to the best-matched frames with the given words. Motivated by this, we design a multi-modal attentive encoder that can see the whole frames and words at once, and attend to the related parts between the two modalities. To this end, the multi-modal attentive encoder is designed with transformer \cite{vaswani2017attention} by stacking the two modal context representations, $f_v$ and $f_t$, in the time dimension to construct multi-modal attention similar to \cite{chen2020uniter,lee2020multimodaltransformer}. Therefore, the multi-modal attentive encoder can model the inter-modal correspondences through attention, which is the key in our visual forced alignment problem. Let $E_m$ be the multi-modal attentive encoder, then modelling inter-modal correspondences can be written as follows,
\begin{align}
    o_m = E_m(f_m), \quad \text{where} \;\; f_m=f_v\oplus f_t,
\end{align}
$f_m\in\mathbb{R}^{(T+S)\times D}$ represents the stacked representations of two modal contexts in the time dimension, $\oplus$ is the concatenation operation, and $o_m\in\mathbb{R}^{(T+S)\times D}$ represents the output features modeled the inter-modal correspondences.

\subsection{Task Augmentation with Anomaly Case Prediction}
Given transcriptions often have errors so there may exist mismatching words with the spoken speech. These errors are induced by restarts during filming, deviations in read speech from the transcription, or incorrect annotations by listeners in spontaneous speech \cite{mcauliffe2017MFA}. These errors should be considered for developing reliable visual forced alignment technology. To this end, we augment the alignment task with anomaly case prediction which can distinguish the mismatch between the transcription and the talking face video. There could be three types of mismatch errors, addition, deletion, and substitution. The addition is the case where an additional word is annotated which is not indeed present in the talking face video. The deletion indicates that a word is not annotated properly so it does not exist in the transcription while it appears in the video. Finally, the substitution is the combination of addition and deletion so the original word spoken in the video is changed with the wrong word in the transcription. The examples for the anomaly cases can be found in Fig. \ref{fig:2}(b). We augment the learning problem to be able to find these three anomaly cases during performing the alignment. To achieve this, we build two predictors, Text Alignment Predictor (TAP) which predicts the time-alignment information with the deletion anomaly case, and Text Presence Predictor (TPP) which predicts the presence of words in the talking face video including the addition anomaly error. Since the substitution error is the combination of deletion and addition, it can be detected when both addition and deletion errors are identified.

\subsubsection{Text Alignment Predictor.}
The output of TAP is designed to have the same frame lengths as the input video. Thus, each output frame can represent word information corresponding to each video frame, and eventually this holds the time-alignment information of the given transcription matched with the input video. Moreover, when a spoken word is not founded in the transcription (\ie, deletion error), TAP is guided to predict a deletion token, $<$d$>$, with its timeline to inform there does not exist the spoken word in the transcription. The learning of TAP can be written as follows,
\begin{align}
    \hat{y}_a=\text{TAP}(o_m^{1:T}),
\end{align}
where $o_m^{1:T}$ represents the first $T$ features of the inter-modal correspondences $o_m$ to be utilized for the alignment prediction and $\hat{y_a}\in\mathbb{R}^{T\times N}$ is the predicted alignment information with $N$ classes. Then, TAP is guided with cross entropy loss as follows,
\begin{align}
    \mathcal{L}_{TAP} = - \sum_{i=1}^{T}y_a^i\log(\hat{y}_a^i),
\end{align}
where $y_a$ is the ground truth alignment. 
For the prediction target, we set it to the position of each word, which enables the model to align the transcription without depending on a predefined word dictionary. The position of each word (\ie, temporal index) is repeated by its actual duration with the same frame rate as the video. Thus, the resulting ground truth alignment is composed of prediction targets with the same number as video frames, similar to \cite{ren2019fastspeech}. Specifically, if the frame duration for each word $\{x_t^1, x_t^2, \dots x_t^S\}$ in transcription $x_t$ is $\{d^1, d^2, \dots, d^S\}$, then the ground truth alignment becomes $\{R(1, d^1), R(2, d^2), \dots, R(S, d^S)\}$, where $R(\alpha, \beta)$ represents $\beta$ times repetition of $\alpha$. For example, when the transcription is composed of 3 words as $\{x_t^1, x_t^2, x_t^3\}$ with duration of 1, 3, and 2 frames each, then the ground truth alignment is $\{1, 2, 2, 2, 3, 3\}$. Please note we additionally use two additional classes for silence (\ie, $<$s$>$) and deletion (\ie, $<$d$>$) where the silence token is inserted for the silence frames and the deletion token is inserted to represent the deletion error case. Therefore, the total class to be predicted by TAP is $S+2$ (\ie, $N=S+2$).

\subsubsection{Text Presence Predictor.}
TPP is for predicting whether the word in the transcription is indeed spoken by the input video. Therefore, it can detect the addition error. The output of TPP has the same length as the number of input words. Thus, each output can hold the presence information of each input word in the video. To guide TPP to correctly predict the presence of input words, it is trained with binary cross entropy loss as follows,
\begin{gather}
    \hat{y}_p=\text{TPP}(o_m^{T+1:T+S}),\\
    \mathcal{L}_{TPP} = - \sum_{i=1}^{S}(y_p^i\log(\hat{y}_p^i) + (1-y_p^i)\log(1-\hat{y}_p^i)),
\end{gather}
where $o_m^{T+1:T+S}$ represents $S$ features after the $T$-th feature of the inter-modal correspondences $o_m$ to be utilized for the presence prediction, $\hat{y_p}\in\mathbb{R}^{S\times 1}$ is the predicted presence, and $y_p$ is the ground truth presence. The ground truth presence is set to 1 when the word appears in the video and 0 for the opposite situation (\ie, addition error). During training, we randomly add an additional word selected from the word dictionary of the training dataset to the transcription and change the corresponding ground truth presence to 0, for simulating the addition anomaly case.

The total objective function, $\mathcal{L}$, for training the proposed DVFA is the summation of the two predefined loss functions, $\mathcal{L} = \mathcal{L}_{TAP} + \mathcal{L}_{TPP}$. The whole model can be trained in an end-to-end manner with the objective function $\mathcal{L}$.

\section{Experiments}
\subsection{Dataset}
To evaluate the performance of the proposed DVFA, we utilize two popular video corpus databases, LRS2 \cite{chung2017lrs2} and LRS3 \cite{afouras2018lrs3}.

{\bf LRS2 dataset} is an English sentence-level video corpus dataset constructed with videos from BBC news programs. It consists of about 224 hours of video from diverse speakers. We utilize the train and test splits of the dataset to develop and test the model. For all frames, the lip region is cropped with a size of 80$\times$80, resized into 112$\times$112, and converted into grayscale, following \cite{hong2022visual}.

{\bf LRS3 dataset} is another large-scale English sentence-level video corpus dataset. The dataset is composed of about 150K utterances collected from the TED and TEDx videos. The train and test splits of the dataset are employed for the experiment. Similar to LRS2, all frames are cropped with a size of 100$\times$100 centered at the lip, resized into 112$\times$112, and converted into grayscale.

For both databases, we utilize Montreal Forced Alignment \cite{mcauliffe2017MFA}, an audio forced alignment method, to generate the ground truth alignments, following \cite{momeni2020KWS,prajwal2021transpotter}. For the data augmentation, the videos are randomly horizontally flipped and their spatial region is randomly erased. During training, to simulate the anomaly situation, the ground truth transcription is perturbed with addition, deletion, and substitution errors with a probability of 10\% each.

\begin{table}[t!]
	\renewcommand{\arraystretch}{1.4}
	\renewcommand{\tabcolsep}{5.4mm}
\resizebox{0.9999\linewidth}{!}{
\begin{tabular}{ccc}
\Xhline{3\arrayrulewidth}
\textbf{Method} & \textbf{MAE} $\downarrow$ & \textbf{ACC} $\uparrow$ \\ \hline
KWS-Net \cite{momeni2020KWS} & 171.5ms & 53.0\% \\
CTC-based \cite{kurzinger2020ctc-align} & 80.4ms & 71.7\% \\ 
Transpotter \cite{prajwal2021transpotter} & 71.7ms & 75.0\% \\ \hline
\textbf{Proposed Method} & \textbf{67.7ms} & \textbf{84.2\%} \\ 
\Xhline{3\arrayrulewidth}
\end{tabular}}
\caption{Alignment performance comparison on LRS2.}
\label{table:1}
\end{table}

\begin{table}[t!]
	\renewcommand{\arraystretch}{1.4}
	\renewcommand{\tabcolsep}{5.4mm}
\resizebox{0.9999\linewidth}{!}{
\begin{tabular}{ccc}
\Xhline{3\arrayrulewidth}
\textbf{Method} & \textbf{MAE} $\downarrow$ & \textbf{ACC} $\uparrow$ \\ \hline
KWS-Net \cite{momeni2020KWS} & 262.9ms & 42.6\% \\
CTC-based \cite{kurzinger2020ctc-align} & 124.5ms & 60.1\% \\ 
Transpotter \cite{prajwal2021transpotter} & 167.3ms & 61.8\% \\ \hline
\textbf{Proposed Method} & \textbf{97.7ms} & \textbf{80.2\%} \\ 
\Xhline{3\arrayrulewidth}
\end{tabular}}
\caption{Alignment performance comparison on LRS3.}
\label{table:2}
\end{table}

\begin{table}[t!]
	\renewcommand{\arraystretch}{1.4}
	\renewcommand{\tabcolsep}{6.4mm}
\centering
\resizebox{0.7\linewidth}{!}{
\begin{tabular}{ccc}
\Xhline{3\arrayrulewidth}
\textbf{Dataset} & \textbf{MAE} $\downarrow$ & \textbf{ACC} $\uparrow$ \\ \hline
LRS2 & 176.5ms & 61.3\% \\
LRS3 & 249.1ms & 57.1\% \\ \hline
\Xhline{3\arrayrulewidth}
\end{tabular}}
\caption{Phoneme-level alignment performance of DVFA.}
\label{table:3}
\end{table}

\begin{figure*}[t!]
	\begin{minipage}[b]{1.0\linewidth}
        \centering
    	\centerline{\includegraphics[width=17cm]{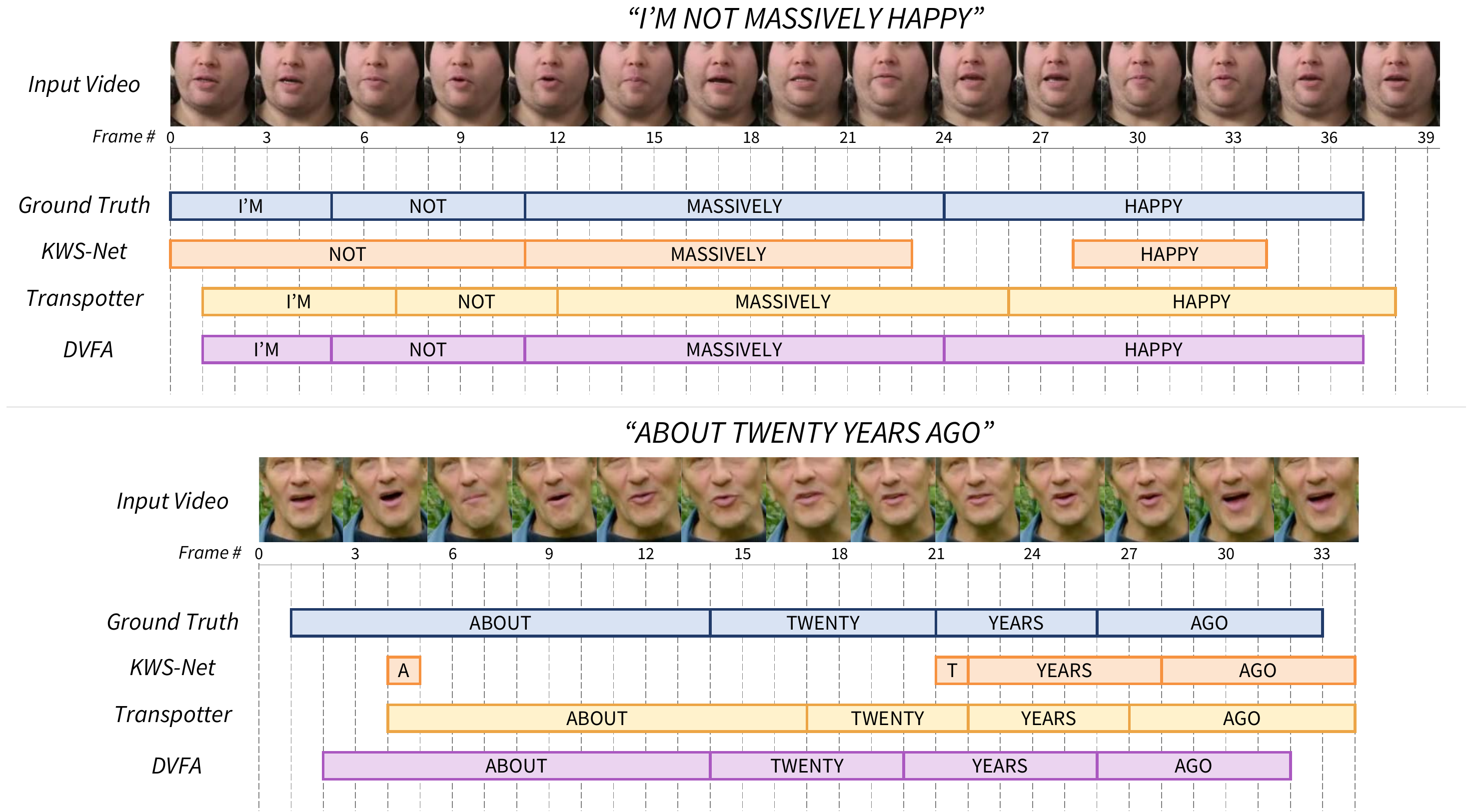}}
    \end{minipage}
	\caption{Qualitative results comparison in transcription-video alignment.}
	\label{fig:3}
\end{figure*}

\subsection{Implementation Details}
For the video encoder, we use a popular architecture in VSR that consists of ResNet-18 \cite{he2016deep} whose first layer is changed with a 3D convolution layer \cite{stafylakis2017reslstm,petridis2018end}. A learnable embedding layer is used for the text encoder. The visual context encoder and the text context encoder are composed with conformer \cite{gulati2020conformer}. The convolution filter size is set to 7, and embedding dimension of 512 (\ie, $D=512$), attention heads of 8, and a total of 4 layers are used for the conformer, in both context encoders. For the text tokenizer, we use sentencepiece \cite{kudo2018sentencepiece}. Since each word is represented with multiple subword tokens, the token-level text representations are merged into word-level representations through average pooling, at the middle of the text context encoders (\ie, after 2nd conformer layer). The multi-modal attentive encoder is designed with 6 layers transformer \cite{vaswani2017attention}. Finally, TAP and TPP are composed of a linear layer, respectively. For training, we use AdamW \cite{kingma2014adam,loshchilov2018decoupled} optimizer, batch size of 32, the initial learning rate of 0.0001, and two NVIDIA RTX A6000 GPUs (48GB).

For the alignment performance measurement, we utilize two metrics, 1) Mean Absolute Error (MAE) which measures the differences between predicted left and right word boundaries and the ground truth left and right boundaries \cite{mcauliffe2017MFA}, and 2) frame-wise prediction accuracy (ACC) which measures whether each video frame is correctly aligned with corresponding word. Since MAE measures the alignment error, the lower value indicates better alignment performance, while higher frame-wise prediction accuracy represents better alignment performance.

\subsection{Experimental Results}
\subsubsection{Transcription-Video Alignment.}
For the baseline methods, we use CTC-based alignment method \cite{kurzinger2020ctc-align} using a VSR model of \cite{ma2021confavsr}, and bring two visual keyword spotting methods, KWS-Net \cite{momeni2020KWS} and Transpotter \cite{prajwal2021transpotter}, which can predict whether the input word exists in the video and align a given word from the video. To align the whole transcription with the video using the keyword spotting methods, we enter every words in the transcription to the models and obtain the entire alignment maps for the sentence. Also, to measure the alignment performance, we assume that the given transcription is perfect so there is no anomaly cases. Therefore, if the models, KWS-Net, Transpotter, and DVFA, predict a given word is not present in the video, its alignment is approximated by using the predicted timelines of front and back words. Table \ref{table:1} shows the visual forced alignment performances on LRS2 dataset. Compared to the state-of-the-art keyword spotting method, Transpotter, and the CTC-based method, the proposed method outperforms in both metrics, MAE and ACC. Especially, the lowest MAE shows the proposed DVFA can well align the transcription with the input talking face video, accurately. Moreover, by achieving the frame-wise accuracy of 84.2\% which surpasses the previous method by 9.2\%, we can confirm the effectiveness of the proposed method in visual forced alignment. Table \ref{table:2} shows comparisons of the alignment performance on LRS3 dataset. We can find consistent results with that of LRS2. Since the proposed DVFA aligns whole sentences at once in contrast to the previous works operating at the word level, it can better align the text and video by considering the context information in the sentence even when the given word is short. The qualitative comparison is shown in Fig. \ref{fig:3}. The results confirm that the proposed DVFA aligns the words accurately by showing the best-matched results with the ground truth alignment. KWS-Net fails to align the word ``I'M'' in the upper example by aligning it to between 27 and 35 frames, which is omitted in the figure due to overlap with the word ``HAPPY''.

In addition, the proposed DVFA can also be utilized to align the text at the phoneme level by changing the word-level inputs into the phoneme-level inputs. To examine the performance of DVFA in aligning at the phoneme level, instead of the word level, we conduct the experiment by tokenizing the input transcription into phonemes. The phoneme-level alignment performance can be found in Table \ref{table:3}. We observe that the phoneme-level alignment performance is overall lower than the word-level alignment performance. This result could be derived from that the video frames are more sparse than the audio so some phonemes remain shorter than one frame duration. For example, one frame in a 25fps video corresponds to 40ms but some phonemes changes within 30ms which cannot be easily detected by watching the video only.

\subsubsection{Detecting Anomaly Errors.}
To validate whether DVFA can distinguish the anomaly errors, we artificially introduce the anomaly cases, addition and deletion errors, to the test set and evaluate the detection accuracy.
Firstly, addition error is introduced to the input transcription with a 50\% probability, so random prediction yields 50\% accuracy. The detection performance for the addition error by TPP achieves 98.0\% and 97.0\% accuracies on LRS2 and LRS3, respectively, and it is shown in Table \ref{table:4}. Similarly, when deletion error is introduced, the detection performance for the deletion error by TAP is 89.0\% and 83.9\% on LRS2 and LRS3. The results confirm that the proposed DVFA can detect the anomaly cases, the mismatches between the input transcription and input talking face video, during performing the forced alignment.

\begin{table}[t!]
	\renewcommand{\arraystretch}{1.4}
	\renewcommand{\tabcolsep}{4.8mm}
\resizebox{0.9999\linewidth}{!}{
\begin{tabular}{cccc}
\Xhline{3\arrayrulewidth}
\textbf{Dataset} & \textbf{Method} & \textbf{Addition} & \textbf{Deletion} \\ \hline
- & Random & 50\% & 50\% \\ \hdashline
LRS2 & \textbf{Proposed Method} & 98.0\% & 89.0\% \\
LRS3 & \textbf{Proposed Method} & 97.0\% & 83.9\% \\ \hline
\Xhline{3\arrayrulewidth}
\end{tabular}}
\caption{Performance of DVFA in detecting anomaly errors.}
\label{table:4}
\end{table}

\begin{table}[t!]
	\renewcommand{\arraystretch}{1.2}
	\renewcommand{\tabcolsep}{5.8mm}
\centering
\resizebox{0.7\linewidth}{!}{
\begin{tabular}{cccc}
\Xhline{3\arrayrulewidth}
\multicolumn{3}{c}{\textbf{Number of Layers}} & \multirow{2}{*}{\textbf{ACC}$\uparrow$} \\ \cline{1-3}
$C_v$ & $C_t$ & $E_m$ &  \\ \hline
8 & 8 & 6 & 82.9\% \\
4 & 8 & 6 & 82.4\% \\ 
8 & 4 & 6 & 82.3\% \\ 
\textbf{4} & \textbf{4} & \textbf{6} & \textbf{84.2\%} \\
4 & 4 & 4 & 83.8\% \\
4 & 4 & 8 & 84.1\% \\ 
\Xhline{3\arrayrulewidth}
\end{tabular}}
\caption{Ablation results according to the number of layers.}
\label{table:5}
\end{table}

\begin{table}[t!]
	\renewcommand{\arraystretch}{1.4}
	\renewcommand{\tabcolsep}{8.5mm}
\centering
\resizebox{0.7\linewidth}{!}{
\begin{tabular}{cc}
\Xhline{3\arrayrulewidth}
\textbf{Prediction target} & \textbf{ACC} $\uparrow$ \\ \hline
Word &  78.1\% \\
\textbf{Position} & \textbf{84.2\%} \\
\Xhline{3\arrayrulewidth}
\end{tabular}}
\caption{Ablation results according to prediction targets.}
\label{table:6}
\end{table}

\subsubsection{Ablation Study.}
We perform an ablation study to examine the effect of the number of layers in each module of visual context encoder, text context encoder, and multi-modal attentive encoder. The ablation results obtained by differing the number of layers are shown in Table \ref{table:5}. We obtain the best frame-wise accuracy when the number of layers in visual context encoder $C_v$, text context encoder $C_t$, and multi-modal attentive encoder $E_m$ are 4, 4, and 6, respectively. In addition, we investigate which kinds of prediction target is more effective. To this end, we set the prediction target to words so the output of TAP is set to classify words instead of their position (\ie, temporal index). The ablation results according to prediction targets are shown in Table \ref{table:6}. Compared to the results obtained using the position of each word as the prediction target (84.2\%), the lower frame-wise accuracy (78.1\%) is obtained by using words as the prediction target. Since predicting words requires more classes to be predicted and needs a predefined word dictionary, it is more difficult, and the model cannot align a new word.

\begin{figure}[t!]
	\begin{minipage}[b]{1.0\linewidth}
		\centering
		\centerline{\includegraphics[width=8.5cm]{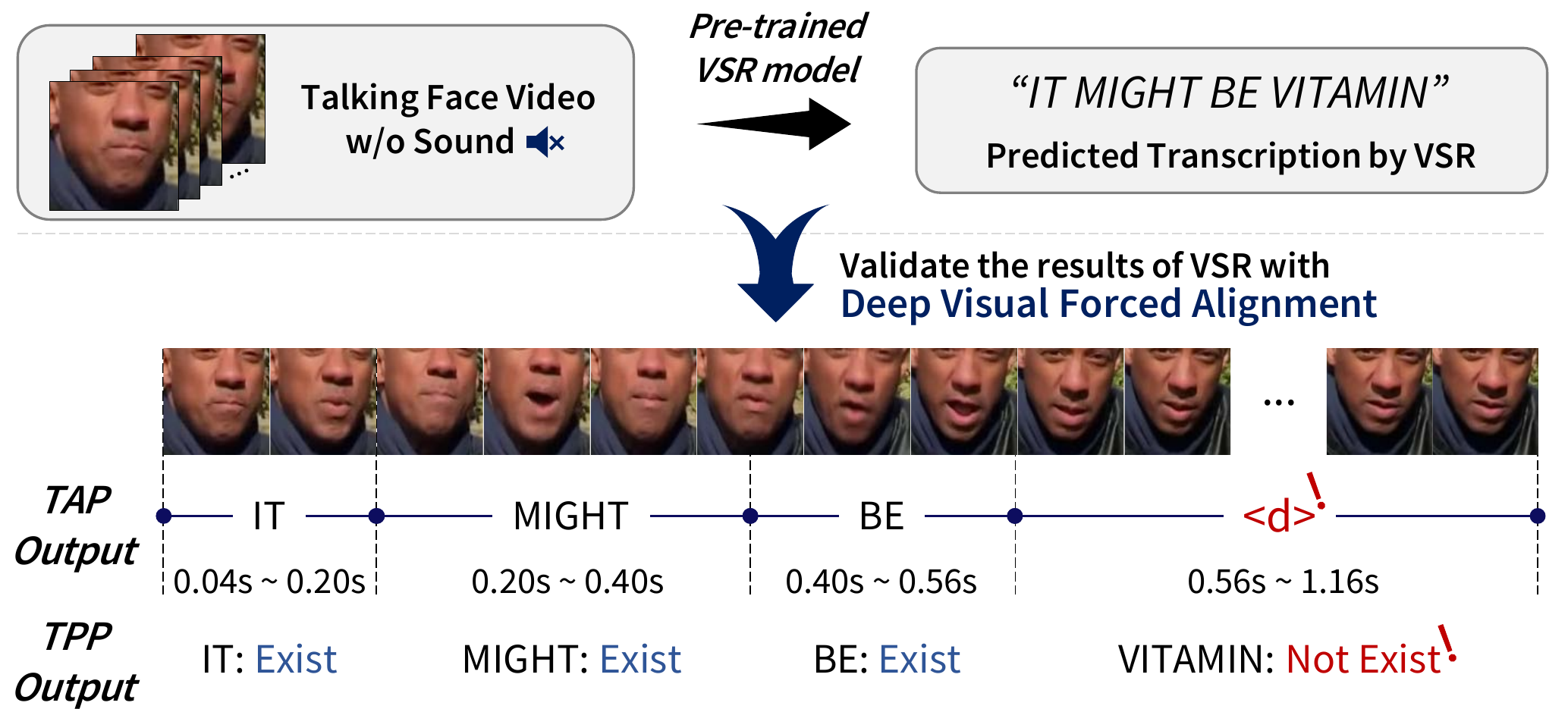}}
	\end{minipage}
	\caption{DVFA can be an interpreter for the outputs of VSR.}
	\label{fig:4}
\end{figure}

\subsubsection{Deep Visual Forced Alignment as an Interpreter.}
As the proposed DVFA can detect the anomaly cases when performing the text-video alignment, it can also be utilized as an interpreter for VSR. Since not all prediction results from current VSR methods are acceptable, there is a requirement to filter only reliable results, and this can be achieved with the proposed DVFA. For example, when we have a speech recognized transcription obtained by a VSR model, DVFA finds the time-alignment map and the mismatches between the predicted transcription and input video. Then, users can find the reliable words from the whole predicted sentence by watching the time-aligned text and video with the anomaly candidates. In order to confirm whether DVFA can be utilized as an interpreter for the VSR, we perform the alignment experiment by using the predicted transcriptions from one of the state-of-the-art VSR models \cite{kim2022distinguishing}. The aligned example is shown in Fig. \ref{fig:4}. The predicted transcription by the VSR model is `IT MIGHT BE VITAMIN'. By putting the predicted transcription into the proposed DVFA with the talking face video, TAP aligns the correctly predicted words with the video while alerting the mismatches between input words and the talking face video. In the figure, the word `VITAMIN' is detected as the deletion case by the TAP and as the addition case by the TPP, which eventually is the substitution case. Therefore, the user can infer that the predicted word `VITAMIN' by the VSR model is a false prediction. Please note that the ground-truth transcription for the video is `IT MIGHT BE VIABLE', which confirms the usability of DVFA as an interpreter.

\section{Conclusion}
In this paper, we proposed a novel Deep Visual Force Alignment (DVFA) for time-aligning the input transcription with the input talking face video without using speech audio. Through the proposed multi-modal attentive module, the frames saying a given word can be retrieved and be time-aligned. Moreover, by augmenting the alignment task with anomaly case detection, DVFA can robustly perform alignment to mismatches between the transcription and the video, and can be utilized to interpret the outputs of VSR models.

\section{Acknowledgments}
This work was supported by the National Research Foundation of Korea (NRF) grant funded by the Korea government (MSIT) (No. NRF-2022R1A2C2005529).

\bibliography{main.bib}
\end{document}